\documentclass{article}
\usepackage{ICLR27_style/iclr2027/iclr2027_conference,times}
\usepackage{amsmath}
\usepackage{amssymb}
\usepackage{booktabs}
\usepackage{array}
\usepackage{float}
\usepackage{wrapfig}
\usepackage{tikz}
\usetikzlibrary{positioning,arrows.meta}
\usepackage{hyperref}
\usepackage{url}

\newif\ifarxiv
\arxivtrue

\title{COREM: Cosine-Relation Momentum Reshaping with Stateful Writeback}

\ifarxiv
\author{
    \normalfont
    \hspace{-0.2em}Yan Wang$^{\dagger}$\thanks{Equal contribution. $\dagger$Corresponding author.}\\
    Aalto University\\
    \texttt{yan.1.wang@aalto.fi}
    \And
    \normalfont\mdseries
    Xiaochuan Wang$^{*}$\\
    Helsinki University\\
    \texttt{xiaochuan.wang@helsinki.fi}
    \And
    \normalfont\mdseries
    Yuxiang Sun\\
    City University of Hong Kong\\
    \texttt{yx.sun@cityu.edu.hk}
}

\iclrfinalcopy
\fi

\begin{document}

\maketitle
\ifarxiv
\fancyhead{}
\fi

\begin{abstract}
Matrix-valued optimizer states may contain relational structure that is not captured by treating their entries independently. We study whether relations within matrix-valued optimizer states can be exploited to improve optimization. To this end, we introduce a unit–relation–transform abstraction and instantiate it as COREM, a Cosine-Relation Momentum Reshaping method with stateful writeback. COREM partitions the momentum state into update units, computes cosine relations among them, and uses these relations to reshape the momentum before writing the transformed state back to the optimizer. This stateful mechanism allows the reshaped momentum to affect not only the current update but also future optimization dynamics.
We evaluate COREM on CIFAR-10 with an MLP and on enwik8 with a Transformer. Compared with Muon, COREM shows lower early-stage step efficiency but stronger improvement in the mid-to-late stages of training, achieving better final validation performance on CIFAR-10 and comparable final performance on enwik8. Spectral diagnostics on enwik8 show that COREM consistently increases entropy effective rank and reduces the concentration of singular energy in dominant modes, while preserving an anisotropic spectrum. For square matrix updates, COREM requires approximately 13.3\% of the transformation FLOPs of Muon with five Newton–Schulz iterations. 
\end{abstract}

\section{Introduction}

Modern optimization methods increasingly exploit structure beyond
coordinate-wise rescaling when transforming parameter updates. Adaptive
methods such as AdaGrad~\citep{duchi2011adagrad} and
Adam~\citep{kingma2015adam} use statistics of past gradients to
modulate updates, while structured approaches such as
K-FAC~\citep{martens2015kfac}, Shampoo~\citep{gupta2018shampoo}, and
SOAP~\citep{vyas2025soap} exploit richer matrix- or tensor-level
structure through curvature approximation or preconditioning. These
developments demonstrate that structure within an optimizer state can
provide useful information for determining how updates are transformed.

A complementary source of structure lies in the relations among
subcomponents of a structured optimizer state, which we refer to as
update units. Explicit relation modeling has proved useful in relational
reasoning~\citep{santoro2017relation},
self-attention~\citep{vaswani2017attention,devlin2019bert,dosovitskiy2021vit,carion2020detr}, and non-local
computation~\citep{wang2018nonlocal}, where interactions among units
condition subsequent aggregation or transformation. This motivates a
central question: can relations among optimizer-state units similarly
be used to determine how the state itself should be transformed?

To study this question, we introduce a
\textbf{unit--relation--transform abstraction},
\[
\boxed{
\text{units}
\;\rightarrow\;
\text{relations}
\;\rightarrow\;
\text{relation-induced transformation}.
}
\]
The abstraction separates three modeling choices: the update units,
the relation function, and the transformation induced by the resulting
relation structure.

We instantiate this abstraction for matrix-valued momentum states as
\textbf{COREM: Cosine-Relation Momentum Reshaping with stateful
writeback}. COREM constructs signed cosine relations among momentum
units and uses the resulting relation operator to reshape the state.
COREM does not impose a fixed target geometry such as exact orthogonality, while its transformation is conditioned on the relational structure of the current momentum state. The transformed
momentum is then written back into the optimizer state, allowing the
reshaped state to influence subsequent momentum accumulation.

Muon~\citep{jordan2024muon} provides a natural reference because it
also transforms matrix-valued momentum, but through an
orthogonalization-oriented mechanism. Across CIFAR-10 and enwik8,
COREM achieves competitive final performance while exhibiting distinct
optimization dynamics and substantially lower transformation cost.
Spectral diagnostics further indicate that COREM redistributes
singular-mode energy while preserving nontrivial spectral anisotropy.

Our contributions are threefold:
\begin{enumerate}
    \item We introduce a \textbf{unit--relation--transform abstraction}
    for constructing optimizer-state transformations from relations
    among update units.

    \item We instantiate this abstraction as \textbf{COREM}, which uses
    signed cosine relations and stateful writeback to reshape
    matrix-valued momentum without enforcing a fixed target geometry.

    \item We evaluate \textbf{COREM} on CIFAR-10 and enwik8, where it achieves
    competitive final performance and exhibits distinct optimization
    dynamics relative to Muon. We further characterize its spectral
    reshaping behavior through optimizer-state diagnostics.
\end{enumerate}
\section{Related Work}

\subsection{Adaptive and Structured Optimizers}

Adaptive methods such as AdaGrad~\citep{duchi2011adagrad} and Adam~\citep{kingma2015adam} use statistics of past gradients to modulate updates, typically through coordinate-wise adaptation. Adafactor~\citep{shazeer2018adafactor} further reduces optimizer-state memory through factored second-moment estimates for matrix-valued parameters.

A complementary line of work exploits matrix- or tensor-level structure. Shampoo~\citep{gupta2018shampoo} uses structured preconditioning across tensor dimensions, while SOAP~\citep{vyas2025soap} combines Shampoo-style preconditioning with Adam-like adaptive updates in an evolving structured basis. These methods move beyond independent scalar coordinates by introducing richer moment or preconditioning structure.

COREM approaches structured optimization from a different direction. Rather than estimating a richer preconditioner, it constructs a transformation from relations among update units within the current optimizer state. Its modeling object is therefore the relational geometry among update units rather than a diagonal or matrix-valued estimate of update scale or curvature.

\subsection{Matrix-Valued Momentum and Spectral Transformations}

Another line of optimizer design directly reshapes matrix-valued updates. Matrix-function methods, including Newton--Schulz iterations, provide practical ways to modify matrix geometry without explicitly computing a full singular value decomposition~\citep{higham2008functions}.

Muon~\citep{jordan2024muon} applies an orthogonalization-oriented Newton--Schulz transformation to matrix-valued momentum. COREM also reshapes matrix-valued momentum, but constructs its transformation from pairwise directional relations rather than from a prescribed orthogonalization objective. Although both can involve cubic terms of the form $XX^\top X$, this similarity is algebraic rather than algorithmic: Newton--Schulz iterations approximate a matrix function associated with orthogonalization, whereas COREM uses a dynamically normalized relation operator to reshape the optimizer state. Muon therefore provides a natural reference point for comparing two distinct mechanisms for transforming matrix-valued momentum.

\subsection{Relation-Based Computation}

Relation-conditioned computation is widely used in representation learning. Relation Networks~\citep{santoro2017relation}, self-attention~\citep{vaswani2017attention,devlin2019bert,dosovitskiy2021vit,carion2020detr}, Graph Attention Networks~\citep{velickovic2018gat}, and non-local neural networks~\citep{wang2018nonlocal} all construct input-dependent interactions among units and use those interactions to control aggregation or transformation.

A broad class of such operations can be written schematically as
\[
X' = R(X)X,
\]
where $R(X)$ is a relation operator constructed from the current input. COREM shares this general relation-conditioned computation pattern, but applies it to optimizer-state units rather than feature representations. Its relation operator is built from signed cosine similarities and enters a residual momentum transformation rather than a softmax-style feature aggregation rule.

The connection to attention and graph-based computation is therefore structural rather than architectural. Taken together, these works motivate COREM as a relation-conditioned optimizer-state transformation that is distinct from both structured preconditioning and orthogonalization-oriented update transforms.

\section{Method}
\label{sec:method}

\subsection{Overview}

COREM transforms a momentum state through dynamically constructed relations among update units. It follows the pipeline
\[
\begin{tikzpicture}[
    >=Stealth,
    every node/.style={font=\small},
    pipelinebox/.style={
        draw=black!45,
        rounded corners=2pt,
        fill=black!3,
        text width=30mm,
        align=center,
        minimum height=8mm,
        inner sep=4pt
    },
    flow/.style={
        ->,
        line width=0.65pt,
        black!70
    }
]

\node[pipelinebox] (units) at (0,0)
    {Unit construction};

\node[pipelinebox] (relations) at (3.8,0)
    {Relation modeling};

\node[pipelinebox] (operator) at (7.6,0)
    {Relation operator};

\node[pipelinebox] (transform) at (7.6,-1.25)
    {State transformation};

\node[pipelinebox] (restore) at (3.8,-1.25)
    {Norm restoration};

\node[pipelinebox] (writeback) at (0,-1.25)
    {State writeback};

\draw[flow] (units.east) -- (relations.west);
\draw[flow] (relations.east) -- (operator.west);

\draw[flow] (operator.south) -- (transform.north);

\draw[flow] (transform.west) -- (restore.east);
\draw[flow] (restore.west) -- (writeback.east);

\end{tikzpicture}
\]
Let $G_t$ be the gradient and $M_{t-1}$ the previously written-back momentum state. The momentum candidate, COREM transformation, and parameter update are
\[
V_t=\mu M_{t-1}+G_t,\qquad M_t=\mathcal{T}_{\mathrm{COREM}}(V_t),
\qquad \theta_t=\theta_{t-1}-\alpha\,M_t.
\]
Crucially, $M_t$ is retained: $V_{t+1}=\mu M_t+G_{t+1}$. Thus the state recurrence is $M_t=\mathcal{T}_{\mathrm{COREM}}(\mu M_{t-1}+G_t)$, coupling spatial relations among units with temporal momentum memory.

\subsection{Unit--Relation--Transform Abstraction}

Given update units $\mathcal U=\{u_1,\ldots,u_n\}$, a generic
unit--relation--transform construction represents each unit as
$z_i=\phi(u_i)$, defines pairwise relations
$R_{ij}=\mathcal R(z_i,z_j)$, and constructs a relation-induced
operator $A=\mathcal A(R)$ that produces
$\widetilde{\mathcal U}=\mathcal T_A(\mathcal U)$.
This separates the choices of update units, relation functions, and
relation-induced transformations.

For a matrix-valued momentum candidate, COREM first chooses a canonical
orientation in which the shorter dimension indexes update units. The
matrix is transposed when necessary so that
\[
V_t=
[v_{t,1}^{\top};\ldots;v_{t,n}^{\top}]
\in\mathbb R^{n\times d},
\qquad n\le d.
\]
For notational simplicity, $V_t$ denotes this oriented representation
throughout the transformation; the original orientation is restored
before the transformed state is written back.

COREM represents each update unit by its normalized direction:
\[
r_{t,i}
=
\max(\|v_{t,i}\|_2,\epsilon),
\qquad
\widehat v_{t,i}
=
\frac{v_{t,i}}{r_{t,i}},
\qquad
\widehat V_t
=
\operatorname{RowNormalize}(V_t).
\]
where $\epsilon>0$ is a small numerical constant used to avoid
division by zero.

\subsection{Cosine-Relation Construction}

COREM constructs signed cosine relations among the normalized update
units and removes self-relations:
\[
C_t
=
\operatorname{OffDiag}(\widehat V_t\widehat V_t^\top) = \widehat V_t\widehat V_t^\top-I,
\qquad
\rho_t
=
\|C_t\|_\infty
=
\max_i\sum_j |(C_t)_{ij}|,
\qquad
\overline C_t
=
\frac{C_t}{\rho_t+\epsilon}.
\]
Thus, $C_t$ retains both positive and negative cross-unit directional
relations, while $\rho_t$ normalizes their maximum absolute row-sum
scale.

Because $C_t$ is symmetric,
$\|C_t\|_2
\le
\sqrt{\|C_t\|_1\|C_t\|_\infty}
=
\rho_t$.
Hence,
$\|\overline C_t\|_2
\le
\rho_t/(\rho_t+\epsilon)
<1$
for $\rho_t>0$ (and $\overline C_t=0$ when $\rho_t=0$), so the
eigenvalues of $\overline C_t$ lie in $(-1,1)$.

\subsection{Relation-Normalized Momentum Reshaping and Spectral Control}
\label{sec:relation-normalized-reshaping}

The normalized momentum state is reshaped by the relation operator:
\[
\widetilde V_t
=
\widehat V_t
-
\eta\overline C_t\widehat V_t
=
(I-\eta\overline C_t)\widehat V_t.
\]

Since the eigenvalues $\lambda_k(\overline C_t)$ of the normalized
relation operator lie in $[-1,1]$, the corresponding relation eigenmode
is scaled by
\[
g_{t,k}
=
1-\eta\lambda_k(\overline C_t)
\in
[1-\eta,\,1+\eta].
\]
Thus, relation normalization bounds the mode-wise effect of the
transformation; in particular, when $0\le\eta<1$, all mode gains remain
positive and no relation eigenmode changes sign.

After relation filtering, COREM restores the Frobenius norm \(\|V_t\|_F\) and returns the transformed momentum to its original orientation, yielding \(M_t\).

\subsection{Connections and Computational Cost}

The relation term has the generic form $R(X)X$, resembling the
relation-conditioned aggregation pattern used in attention-like
operations. COREM instantiates this pattern differently: its relation
matrix is signed and non-softmax, and the relation term enters through
the residual transformation $X'=X-\eta R(X)X$. Ignoring the numerical safeguard, relation normalization, and norm
restoration, $C=\widehat V\widehat V^\top-I$ gives
\[
\widetilde V
=
(1+\eta)\widehat V
-
\eta\widehat V\widehat V^\top\widehat V.
\]
At $\eta=\tfrac12$, this has the cubic form of a Newton--Schulz polar
iteration. This connection is algebraic only: the full COREM method
normalizes the relation operator, restores the Frobenius norm,
and writes the transformed state back into the momentum dynamics.

For $V\in\mathbb R^{n\times d}$, the two dominant matrix products $\widehat V\widehat V^\top$ and $\overline C\widehat V$ cost approximately $4n^2d$ FLOPs. For a square state, this is $4n^3$ FLOPs. A quintic Newton--Schulz iteration uses three dominant products, or approximately $6n^3$ FLOPs; five iterations therefore cost approximately $30n^3$. Hence, for square states, one COREM transform requires $4/30\approx13.3\%$ of the transformation FLOPs of a five-iteration Muon Newton--Schulz transform~\citep{jordan2024muon}.


\section{Experiments}

We evaluate COREM on two complementary benchmarks, CIFAR-10 image classification~\citep{krizhevsky2009cifar} and enwik8 character-level language modeling~\citep{hutter2006enwik8}, and compare it with Muon~\citep{jordan2024muon}. 
Non-matrix parameters are optimized using SGD with momentum.

\subsection{Experimental Setup}

For CIFAR-10, we train a two-hidden-layer MLP with hidden dimensions $256\times256$ and compare COREM against Muon. All CIFAR-10 experiments are run on a MacBook using the Apple MPS backend.
For enwik8, we train a decoder-only Transformer~\citep{vaswani2017attention} with hidden dimension $192$, four layers, six attention heads, and a context length of $256$ for $64{,}000$ optimization steps. COREM and Muon are evaluated under the same model and data configuration. All enwik8 experiments are run on an NVIDIA GeForce RTX 3080 Ti.

\subsection{Main Results}
\label{sec:main-results}

\begin{table}[t]
\centering
\caption{CIFAR-10 validation results over three random seeds. Values denote mean $\pm$ sample standard deviation.}
\label{tab:exp-cifar}
\small
\begin{tabular}{lrrrr}
\toprule
Method & Final Val. Acc. & Best Val. Acc. & Final Val. Loss & Best Val. Loss \\
\midrule
Muon & $59.74\pm1.01\%$ & $60.67\pm0.44\%$ & $1.2157\pm0.0322$ & $1.1713\pm0.0131$ \\
COREM & $61.36\pm0.19\%$ & $61.87\pm0.23\%$ & $1.1450\pm0.0084$ & $1.1282\pm0.0040$ \\
\bottomrule
\end{tabular}
\end{table}

\begin{table}[t]
\centering
\caption{enwik8 validation results after $64{,}000$ steps over three random seeds. Values denote mean $\pm$ sample standard deviation.}
\label{tab:exp-enwik8}
\small
\begin{tabular}{lrrr}
\toprule
Method & Final Val. Loss & Final Val. BPB & Final Val. Accuracy \\
\midrule
Muon & $1.095263\pm0.004475$ & $1.580130\pm0.006457$ & $0.678172\pm0.001344$ \\
COREM & $1.096513\pm0.002470$ & $1.581933\pm0.003563$ & $0.678525\pm0.001018$ \\
\bottomrule
\end{tabular}
\end{table}

According to Table~\ref{tab:exp-cifar}, COREM achieves higher final and
best validation accuracy on CIFAR-10, together with lower final and best
validation loss and lower cross-seed variation. Table~\ref{tab:exp-enwik8},
in contrast, shows comparable final performance on enwik8: Muon obtains
slightly lower validation loss and BPB, whereas COREM attains slightly
higher validation accuracy with lower cross-seed variability.

\subsection{Optimization Dynamics}

\begin{figure}[!t]
    \centering
    \includegraphics[width=\linewidth]{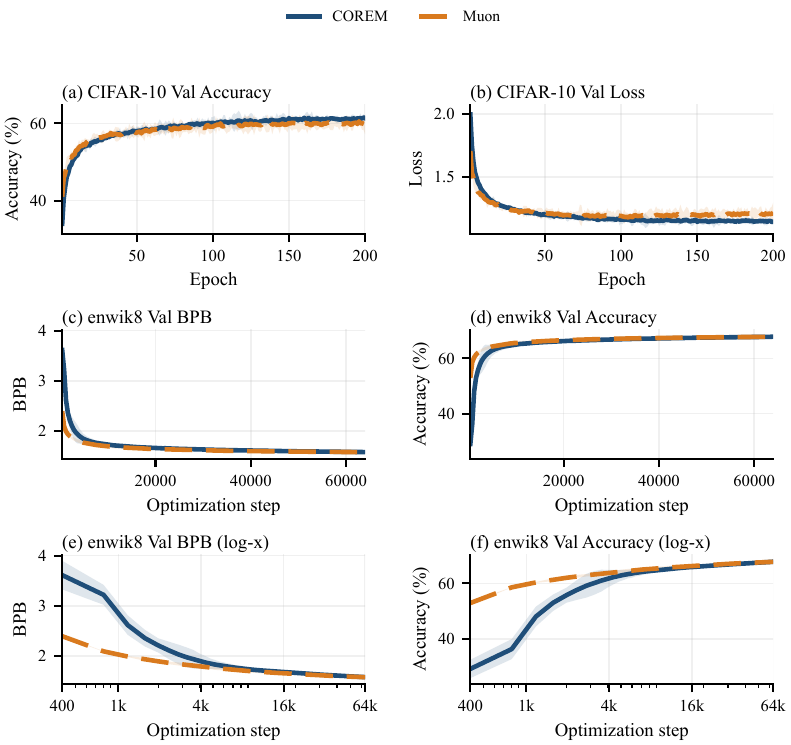}
    \caption{Validation trajectories for COREM and Muon on CIFAR-10 and
    enwik8. The first two rows use linear optimization time, while the
    third row shows enwik8 trajectories with logarithmic optimization
    steps. Shaded regions denote the three-seed uncertainty bands.}
    \label{fig:optimization-dynamics}
\end{figure}

The final metrics in Section~\ref{sec:main-results} do not fully reflect the
different optimization trajectories of COREM and Muon. As shown in
Figure~\ref{fig:optimization-dynamics}, Muon progresses faster early in
training, whereas COREM exhibits relatively stronger late-stage
improvement.

\subsubsection{CIFAR-10}

On CIFAR-10, Muon reaches $50\%$ and $55\%$ validation accuracy after
approximately $5.3$ and $16.3$ epochs on average, compared with $8.0$
and $20.7$ epochs for COREM. In contrast, over the final $25$ epochs,
COREM averages $61.20\%$ validation accuracy, compared with $60.00\%$
for Muon. Thus, Muon reaches moderate accuracy levels earlier, while
COREM attains stronger late-stage validation performance.

\subsubsection{enwik8}

A similar pattern appears on enwik8. Muon reaches validation loss
$\leq1.2$ after approximately $7.2\mathrm{k}$ steps, compared with
$9.7\mathrm{k}$ for COREM, and reaches BPB $\leq1.6$ after
$40.7\mathrm{k}$ versus $48.4\mathrm{k}$ steps. Between
$50\mathrm{k}$ and $64\mathrm{k}$ steps, however, COREM reduces mean
validation BPB from $1.600102$ to $1.581933$, compared with
$1.592999$ to $1.580130$ for Muon, substantially narrowing the
remaining gap by the end of training.

\subsection{Ablation Studies}
\subsubsection{Stateful Momentum Writeback}

We evaluate the role of stateful momentum writeback on CIFAR-10 by
comparing standard COREM with a no-writeback variant, in which the
transformed momentum is used for the current parameter update but the
untransformed momentum is retained for subsequent accumulation. Both
variants are trained for $200$ epochs over three random seeds using
independently tuned operating points.

\begin{wrapfigure}[18]{r}{0.49\textwidth}
    \vspace{-0.5\baselineskip}
    \centering
    \includegraphics[width=\linewidth]{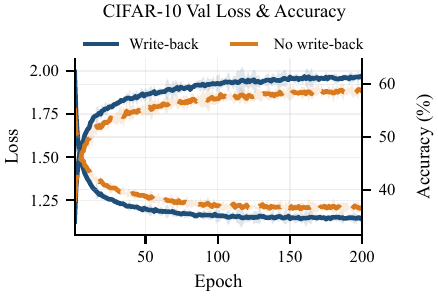}
    \caption{CIFAR-10 validation trajectories for the tuned writeback and
    no-writeback variants. Shaded regions denote the three-seed uncertainty
    bands.}
    \label{fig:writeback-trajectories}
    \vspace{-0.5\baselineskip}
\end{wrapfigure}

As shown in Table~\ref{tab:exp-writeback}, stateful writeback improves
both final and best validation performance. The corresponding
trajectories in Figure~\ref{fig:writeback-trajectories} show that the
writeback variant also reaches the higher-performance regime earlier
and maintains a stronger late-stage validation level.
Because the two variants use independently tuned configurations
($\alpha=0.01,\eta=1.2$ with writeback and
$\alpha=0.0125,\eta=1.0$ without writeback), the comparison should be
interpreted as one between their tuned operating points rather than as a
strict single-hyperparameter causal isolation. Nevertheless, the
no-writeback variant does not recover the performance of stateful COREM
after retuning, supporting writeback as an important component of the
method.


\begin{table}[t]
\centering
\caption{Effect of stateful momentum writeback on CIFAR-10 over three random seeds. Values denote mean $\pm$ sample standard deviation.}
\label{tab:exp-writeback}
\small
\begin{tabular}{lrrrr}
\toprule
Variant & Final Acc. & Best Acc. & Final Loss & Best Loss \\
\midrule
No writeback & $58.61\pm0.44\%$ & $59.47\pm0.13\%$ & $1.2130\pm0.0027$ & $1.1859\pm0.0035$ \\
Writeback & $61.36\pm0.19\%$ & $61.87\pm0.23\%$ & $1.1450\pm0.0084$ & $1.1282\pm0.0040$ \\
\bottomrule
\end{tabular}
\end{table}

\subsubsection{Relation Normalization}

\begin{figure}[H]
    \centering
    \includegraphics[width=\linewidth]{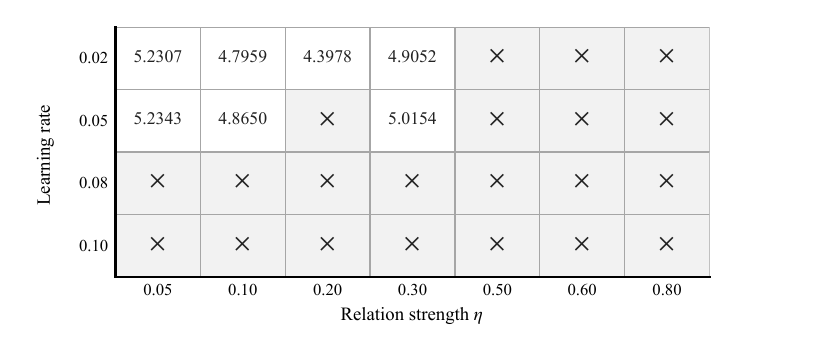}
    \caption{Stability map of the unnormalized COREM variant on enwik8 over
    a 4k-step learning-rate/relation-strength sweep. Each finite configuration
    is annotated with its final validation BPB, while $\times$ denotes a
    numerically unstable or diverged run.}
    \label{fig:no-scale-lr-eta}
\end{figure}

We examine the role of relation-operator normalization by removing the
max-row-$\ell_1$ scaling and performing a $4\mathrm{k}$-step enwik8 sweep
over
\[
\alpha\in\{0.02,0.05,0.08,0.10\},
\qquad
\eta\in\{0.05,0.10,0.20,0.30,0.50,0.60,0.80\}.
\]
As shown in Figure~\ref{fig:no-scale-lr-eta}, only $7$ of the $28$
tested configurations remain numerically stable. In particular, all
configurations with $\alpha\geq0.08$ or $\eta\geq0.50$ become
unstable.

Among the finite runs, the best final validation BPB is $4.3978$,
compared with $1.8556$ for the normalized reference ($\alpha=0.10,\eta=0.50$). Because the two
variants are evaluated over different hyperparameter grids, this
comparison is used primarily to characterize their stable operating
regions rather than as a pointwise matched ablation. The result is
consistent with the spectral-control motivation for relation-operator
normalization developed in Section~\ref{sec:relation-normalized-reshaping}.

\subsection{Analysis of Relation and Spectral Dynamics}

We next examine how COREM reshapes the geometry of the optimizer state during training. The analysis uses diagnostic snapshots from the 64k-step enwik8 run at nine checkpoints, $400$, $800$, $2\mathrm{k}$, $4\mathrm{k}$, $8\mathrm{k}$, $16\mathrm{k}$, $32\mathrm{k}$, $49.6\mathrm{k}$, and $64\mathrm{k}$, covering four Transformer layers and six matrix-valued parameter groups. For each layer-step pair, we record the instantaneous gradient $G_t$, the pre-transformation momentum $V_t=\mu M_{t-1}+G_t$, and the transformed momentum $M_t=\mathcal{T}_{\mathrm{COREM}}(V_t)$. Because writeback is enabled, $V_t$ already contains the previously transformed state $M_{t-1}$; it should therefore be interpreted as an inherited stateful momentum geometry rather than a COREM-free baseline.

\subsubsection{Relation-Scale Heterogeneity}

We examine the raw relation scale
$\rho_t=\|C_t\|_\infty$
throughout training.
\begin{figure}[!t]
    \centering
    \includegraphics[width=\linewidth]{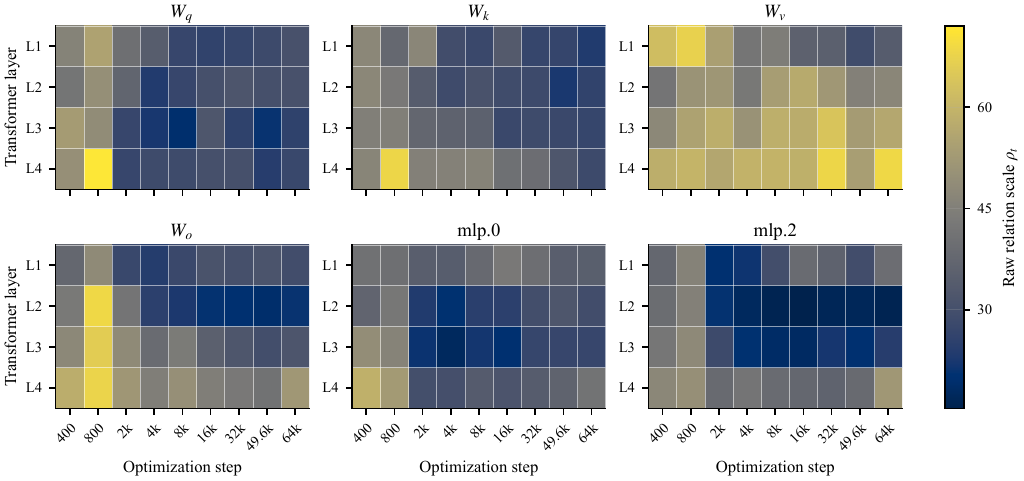}
    \caption{Raw relation-scale dynamics across parameter groups, Transformer layers, and sampled optimization steps on enwik8. Each panel corresponds to one matrix-valued parameter group; rows denote Transformer layers, columns denote sampled optimization steps, and color indicates $\rho_t=\|C_t\|_\infty$. $W_q$, $W_k$, $W_v$, and $W_o$ denote the attention projection matrices, while \texttt{mlp.0} and \texttt{mlp.2} denote the first and second linear projections in the Transformer MLP block. All panels share the same color scale.}
    \label{fig:relation-scale-heatmaps}
\end{figure}
Across the 216 observations, $\rho_t$ varies substantially across
parameter groups, Transformer layers, and optimization steps, ranging
from approximately $15.42$ to $72.02$. As shown in
Figure~\ref{fig:relation-scale-heatmaps}, $W_v$ generally exhibits
larger relation scales, while the MLP projections tend to operate at
lower levels. This variation is consistent with the reduced stability
observed in the unnormalized ablation: without max-row-$\ell_1$ scaling,
a fixed $\eta$ acts on relation operators with substantially different
raw scales.

\subsubsection{Stateful Spectral Maintenance}

We characterize spectral structure using entropy effective rank,
top-10 singular-energy concentration, and a robust 10--90 spectral
condition ratio:
\[
\begin{aligned}
&r_{\mathrm{eff}}(X)
 =
\exp\left(-\sum_i p_i\log p_i\right),
\quad
p_i=\frac{\sigma_i^2(X)}{\sum_j\sigma_j^2(X)},
\\[4pt]
&C_{10}(X)
 =
\frac{\sum_{i=1}^{10}\sigma_i^2(X)}
{\sum_j\sigma_j^2(X)},
\\[4pt]
&\kappa_{\mathrm{robust}}(X)
 =
\frac{\sigma_{\lceil0.1r\rceil}(X)}
{\sigma_{\lceil0.9r\rceil}(X)+\epsilon}.
\end{aligned}
\]
Here, $r$ denotes the number of singular values of $X$.

\begin{figure}[!t]
    \centering
    \includegraphics[width=\linewidth]{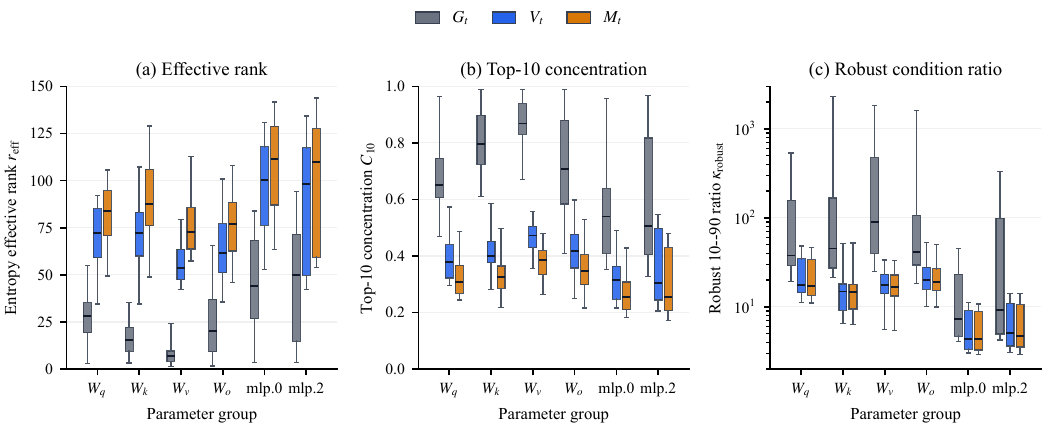}
    \caption{Spectral diagnostics across optimizer states and parameter
    groups on enwik8. Each panel compares $G_t$, $V_t$, and $M_t$
    over the 36 layer--checkpoint observations for each parameter group.
    Panels show entropy effective rank, top-10 singular-energy
    concentration, and the robust 10--90 spectral condition ratio.
    Boxes denote medians and interquartile ranges, with whiskers spanning
    the observed range; the condition-ratio panel uses a logarithmic
    y-axis.}
    \label{fig:spectral-state-diagnostics}
\end{figure}

Across all 216 paired observations, transformed momentum $M_t$
exhibits higher effective rank and lower top-10 singular-energy
concentration than the pre-transformation state $V_t$. In contrast,
the robust condition ratio changes comparatively little from $V_t$
to $M_t$, while both states exhibit substantially lower robust
condition ratios than the instantaneous gradient $G_t$.

Because $V_t=\mu M_{t-1}+G_t$, its spectral geometry already contains
previously transformed momentum propagated through stateful writeback
and should not be interpreted as a COREM-free baseline. The observed
pattern is therefore consistent with stateful propagation of the
reshaped momentum, together with additional current-step spectral
redistribution from $V_t$ to $M_t$.

\section{Conclusion}

We introduced a unit–relation–transform abstraction for optimizer-state transformation and instantiated it as COREM, a Cosine-Relation Momentum Reshaping method with stateful writeback for matrix-valued momentum states. COREM derives its transformation from relations among update units rather than from a prescribed target geometry, and integrates the resulting reshaped state into subsequent momentum dynamics.
Across the evaluated image and language modeling settings, COREM remains competitive with Muon while exhibiting distinct optimization dynamics. The ablation results support the roles of stateful writeback and relation-operator normalization in the proposed design. These findings provide an initial validation of relation-conditioned optimizer-state transformation and motivate future exploration of alternative update units, relation functions, and transformation rules beyond the current matrix-valued setting.

\bibliographystyle{ICLR27_style/iclr2027/iclr2027_conference}
\bibliography{corem}

\end{document}